\pdfoutput=1 
\documentclass[11pt,a4paper]{article}
\usepackage{hyperref}
\usepackage[hyperref]{acl2019}
\usepackage{times}
\usepackage{latexsym}
\usepackage{graphicx}
\usepackage{amsmath}
\usepackage{enumitem}
\usepackage{booktabs}
\usepackage{amssymb}
\usepackage[toc,page]{appendix}

\aclfinalcopy 

\newcommand{\pretty}[1]{\textsc{#1}}
\newcommand{\h}[0]{\mathbf{h}}
\newcommand{\specialcell}[2][l]{%
  \begin{tabular}[#1]{@{}l@{}}#2\end{tabular}}

\title{\pretty{Casa-nlu}: Context-Aware Self-Attentive Natural Language Understanding for Task-Oriented Chatbots}

\author{Arshit Gupta$^\dagger$  
  \quad Peng Zhang$^\ddagger$ \quad Garima Lalwani$^\ddagger$ \quad Mona Diab$^\dagger$\\
$^\dagger$Amazon AI, Seattle \quad $^\ddagger$Amazon AI, East Palo Alto\\
{\tt \{arshig, pezha, glalwani, diabmona\}@amazon.com}}

\date{}

\begin{document}
\maketitle
\begin{abstract}
Natural Language Understanding (NLU) is a core component of dialog systems. It typically involves two tasks - intent classification (IC) and slot labeling (SL), which are then followed by a dialogue management (DM) component. 
Such NLU systems cater to
utterances in isolation, thus pushing the problem of context management to DM. 
However, contextual information is critical to the correct prediction of intents and slots in a conversation. 
Prior work on contextual NLU has been limited in terms of the types of contextual signals used and the understanding of their impact on the model. 
In this work, we propose a context-aware self-attentive NLU (\pretty{Casa-nlu}) model that uses multiple signals, such as previous intents, slots, dialog acts and utterances over a variable context window, in addition to the current user utterance. \pretty{Casa-nlu} outperforms a recurrent contextual NLU baseline on two conversational datasets, yielding a gain of up to 7\% on the IC task for one of the datasets. Moreover, a non-contextual variant of \pretty{Casa-nlu} achieves state-of-the-art performance for IC task on standard public datasets -  \pretty{Snips} and \pretty{ATIS}.
\end{abstract}

\section{Introduction}
\label{sec:introduction_f}

With the advent of smart conversational agents such as Amazon Alexa, Google Assistant, etc., dialogue systems are becoming ubiquitous. 
In the context of enterprises, the majority of these systems target task oriented dialogues with the user trying to achieve a goal, e.g. booking flight tickets or ordering food. \textcolor{black}{Natural Language Understanding (NLU) captures the semantic meaning of a user's utterance within each dialogue turn, by identifying \textbf{intents} and \textbf{slots}. }
An intent specifies the goal underlying the expressed utterance while slots are additional parameters for these intents. These tasks are typically articulated as intent classification (\pretty{IC}) coupled with
sequence tagging task of slot labeling (\pretty{SL}).

\begin{figure}[t]
\centering
\includegraphics[width=150pt, height=120pt, width=0.5\textwidth]{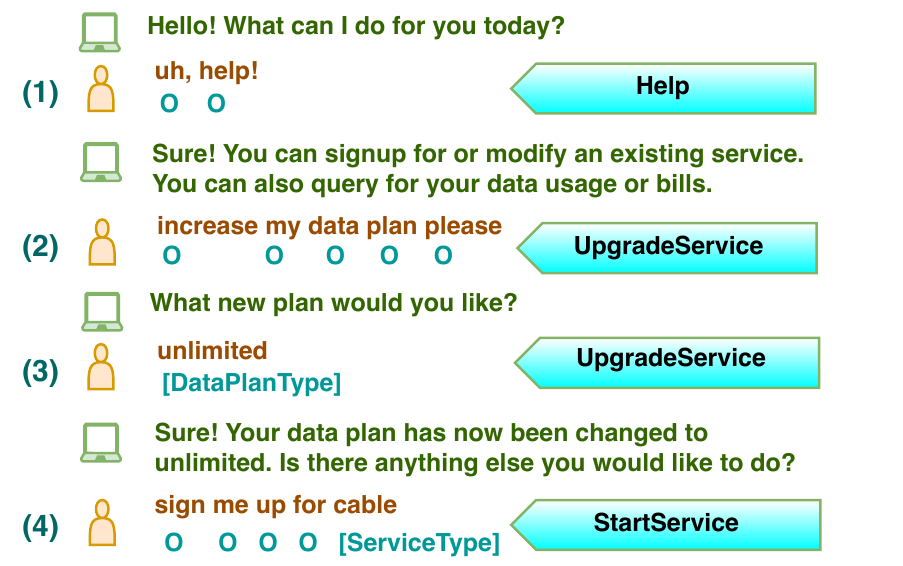}
\vspace{-20pt}
\caption{\label{tease} \small Snippet of a sample Cable data conversation with intents shown in boxes and slots aligned with user request}
\vspace{-20pt}
\end{figure}

Over time, human-machine interactions have become more complex with greater reliance on contextual cues for utterance understanding 
(Figure \ref{tease}).
With traditional NLU frameworks, the resolution of contextual utterances is typically addressed in the DM component of the system using rule-based dialogue state trackers (\pretty{DST}). However, this pushes the problem of context resolution further down the dialogue pipeline, and despite the appeal of modularity in design, it opens the door for significant cascade of errors. To avoid this, end-to-end dialogue systems have been proposed \cite{wen_end2end, end2end_goal}, but, to date, such systems are not scalable in industrial settings, and tend to be opaque where a level of transparency is needed, for instance, to understand various dialogue policies. 

To address 
the propagation of error while maintaining a modular 
framework, \citet{Shi2015ContextualSL} 
proposed adding contextual signals to the joint IC-SL task. However, \textcolor{black}{the contributions of }their work were limited in terms of number of signals and how they were used, rendering the contextualization process still less interpretable. In this work, we present a multi-dimensional self-attention based contextual NLU model that overcomes prior work's shortcomings by supporting variable number of contextual signals (previous utterances, dialogue acts\footnote{Dialog Act signifies the actions taken by the agent such as Close (when an intent is fulfilled), ElicitSlot, etc}, intents and slot labels) that can be used concurrently over variable length of conversation context.
In addition, our model allows for easy visualization and debugging of contextual signals which are essential, especially in production dialogue systems, where interpretability is a desirable feature. Our contributions are:
\begin{itemize}[noitemsep,topsep=0pt,parsep=0pt,partopsep=0pt]
\item Context-Aware Self-Attentive NLU (\pretty{Casa-NLU}) model that uses various contextual signals to perform joint IC-SL task, outperforming contextual and non-contextual NLU baselines by significant margins on two in-house conversational IC-SL datasets
\item Propose a novel non-contextual variant of \pretty{Casa-nlu} that achieves SOTA performance on IC task for both \pretty{Snips} and \pretty{ATIS} datasets
\item Analysis of the various contextual signals' contributions to model performance
\end{itemize}

\section{Related Work}
\label{sec:related_work_f}

There have been numerous advancements in NLU systems for dialogues over the past two decades. While the traditional approaches used handcrafted features and word n-gram based features fed to
SVM, logistic regression, etc. for IC task and conditional random fields (CRF) for SL task \cite{triangular_crf, wang12002combination, raymondAndRiccardi2007}, more recent approaches rely on deep neural networks to jointly model IC and SL tasks \cite{yao2014spokenLU, yao2014recurrentCR, joint_slu_recursive, zhang2016ajm, liu2016jointOS, Goo2018SlotGatedMF}. Attention as introduced by \citet{bahdanau2014neural} has played a major role in many of these systems \cite{liu2016attentionBasedRN, ma2017jointlyTS, li2018ASM, Goo2018SlotGatedMF}, for instance, for modeling interaction between intents and slots in \cite{Goo2018SlotGatedMF}.



\citet{dahlback1989empirical} and \citet{bertomeu2006contextual} studied contextual phenomena and thematic relations in
natural language,
thereby highlighting the importance of using context. 
Few previous works focused on modeling turn-level predictions as \pretty{DST} task \cite{williams2013dialog}. However, these systems 
predict the possible slot-value pairs at utterance level \cite{glad}, making it necessary to maintain ontology of all possible slot values, which is infeasible for certain slot types (e.g., restaurant names). In industry settings, where IC-SL task is predominant, there is also an additional effort involved to invest in rules for converting utterance level dialog state annotations to token level annotations required for SL. 
Hence, our work mainly focuses on the IC-SL task which eliminates the need for maintaining any ontology or such handcrafted rules.

\citet{Bhargava2013EasyCI} used previous intents and slots for IC and SL models. They were followed by \citet{Shi2015ContextualSL} who exploited previous intents and domain predictions to train a joint IC-SL model. However, both these studies lacked comprehensive context modeling framework that allows multiple contextual signals to be used together over a variable context window. 
Also, an intuitive interpretation of the impact of contextual signals on IC-SL task was missing.


\section{\pretty{Casa-nlu}: Context-Aware Self-Attentive NLU}
\label{sec:cdisan_f}
\begin{figure}[ht]
\centering
\includegraphics[width=200pt, height=250pt, width=0.5\textwidth]{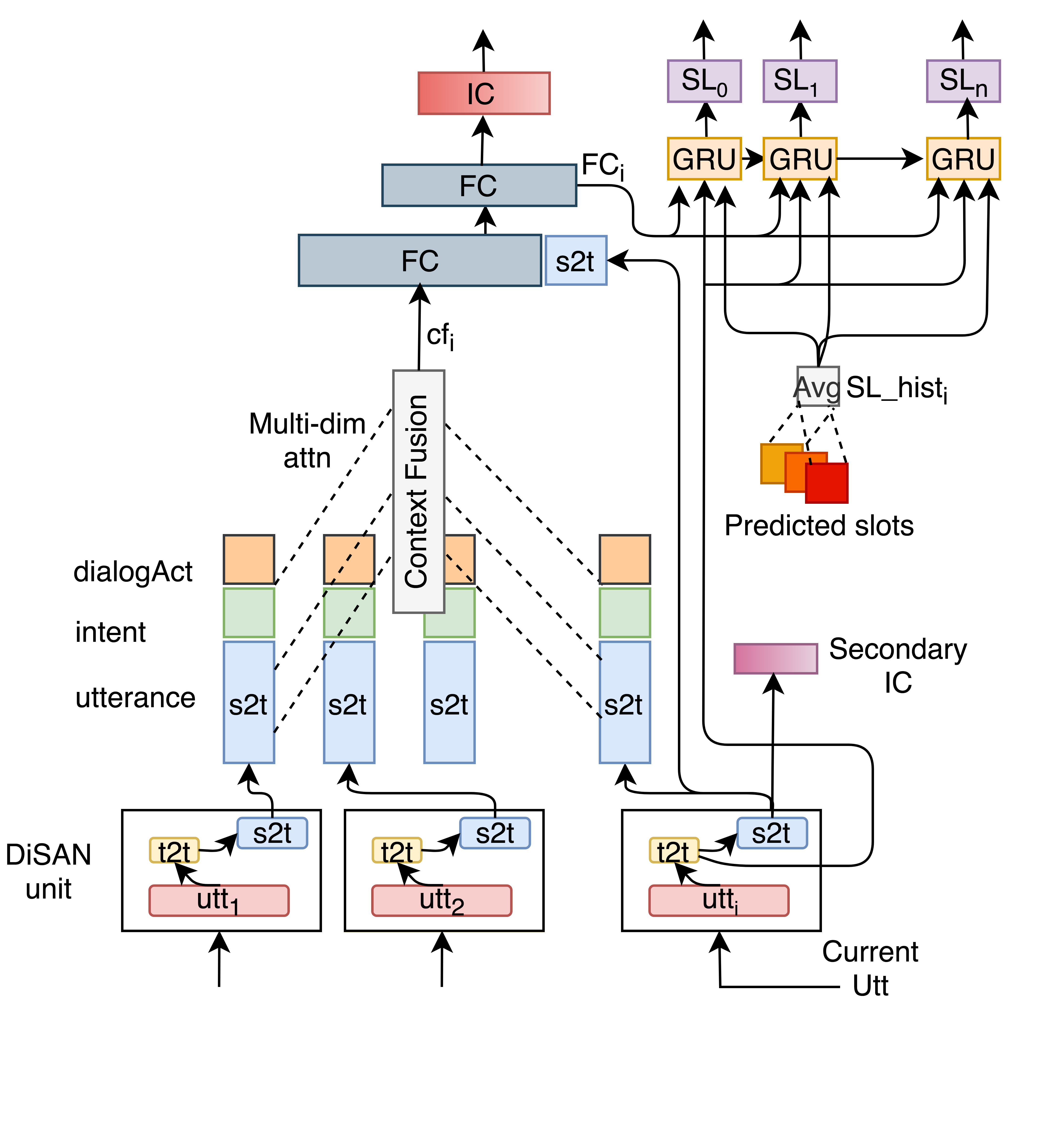}
\vspace{-50pt}
\caption{\label{cSA} \small \pretty{Casa-nlu} model architecture for joint IC-SL}
\end{figure}

Our model architecture is composed of three sub sections - signal encoding, context fusion and IC-SL predictions (Figure \ref{cSA}).
\vspace{-10pt}
\vspace{-10pt}

\subsection{Signal Encoding}

\textbf{Utterance (Utt):} For the utterance encoding, we adopt the directional self-attention \cite{shen2017disan} based encoder (\pretty{DiSAN}), adding absolute position embedding \cite{gehring2017convolutional} to further improve the encoding performance.\footnote{Details provided in Appendix A}
\pretty{DiSAN} unit consists of two types of multi-dimensional attention - word level token2token (t2t) attention followed by sentence level source2token (s2t) attention. For turn $i$, the output of this unit is given by $\h({\text{Utt}}_i) \in\mathbb{R}^{2d_h\times 1}$, 
where $d_h$ is hidden layer size of the \pretty{DiSAN} unit (Figure \ref{cSA}).

\noindent \textbf{Intent / Dialog Act (DA) / Slot Label 
History:} We pass the one-hot representation of intent ground truth labels through an embedding layer to get intent history representation $\h({\text{I}_i}) \in\mathbb{R}^{d_{I}\times 1}$ for any previous turn $i$ with $d_I$ being the intent embedding dimension. Similarly, for DA history, $\h({\text{DA}_i}) \in\mathbb{R}^{d_{DA}\times 1}$. We use a special dummy symbol for the intent and DA of the current turn.  
For slot label history, for turn $i$, we take the average of all slot embeddings that were observed in previous turns giving $\h({\text{SL}\_{\text{hist}}_i}) \in \mathbb{R}^{d_{SL}\times 1}$.
\subsection{Context Fusion}
We combine the vectorized signals together in both spacial and temporal dimensions. For the former, we simply concatenate the contextual signals to get current turn feature vector, i.e. ${\mathbf{T}_i} = [\h({\text{Utt}_i});\h({\text{I}_i}); \h({\text{DA}_i})]$. However, for the latter, concatenation 
becomes intractable if context window is large. To address this issue and automatically learn more relevant components of context for each turn, we 
add a source2token \cite{shen2017disan} multi-dimensional self-attention layer over the turn vectors. This is essentially a per dimension learned weighted average ($\mathbf{cf}_i$) over all the turn vectors within a context window (Equation \ref{eq:attn}).
As shown later in Section \ref{sec:results_f}, this enhances the  model's robustness by learning different attention weights for different contextual signals.
\vspace{-10pt}
\begin{equation}
    \textbf{cf}_i = \sum_{t=i-K}^{i} \textbf{P}(\textbf{T}_t) \odot {\textbf{T}_t} \label{eq:attn}
\end{equation}
\vspace{-10pt}

\noindent where, $K$ (= 3 in our experiments) is the context window, and $\mathbf{T}_t$ and $\mathbf{P}(\mathbf{T}_t)$ are the turn vector and attention weights for $t^{th}$ time step 
respectively. We use padding tokens for $i-K<0$.

One of the shortcomings with such attention mechanism is that it is position invariant. To address this problem, we add learned absolute position embedding $\textbf{p}_c$ 
to the turn matrix $\mathbf{T}$ that learns temporal information across the turns.

\subsection{IC-SL Predictions}
Following \cite{joint_ic_sl, li2018self}, we  train a joint IC-SL model. To improve IC performance for our deep network, we also add a secondary IC loss function, $\text{L}_{\text{Sec}\_\text{IC}}$ at the utterance level (Figure \ref{cSA}). The new aggregated loss is:
\begin{equation}
    \text{L}=\text{L}_{\text{IC}} + \alpha \times \text{L}_{\text{SL}} + \beta \times \text{L}_{\text{Sec}\_\text{IC}}
\end{equation}
\textbf{IC:} At turn $i$, we take the output of context fusion layer $\mathbf{cf}_i$, pass it through a fully connected layer and concatenate the output with the current utterance encoding $\h({\text{Utt}_i})$. 
This is then further projected down using a fully connected layer ($\text{FC}_i$) and finally fed into  \emph{softmax} layer to predict the intent.
\newline


\noindent\textbf{SL:} For turn $i$, the t2t attention output is first fused with the utterance embedding using a fusion gate \cite{hochreiter1997long} to generate $\h_{ij}$ where $j$ represents token index in the utterance. 
Then, for each token position in the utterance,
we apply a sliding window, $w$ (=3) over neighboring words
that transforms each token's embedding space from $\h_{ij}$ to $w \times \h_{ij}$ (not shown in the Figure \ref{cSA}). 
To add contextual information to SL task, each token's dimension is augmented using slot history ($\text{SL}\_\text{hist}_{i}$)  as well as penultimate fully connected layer for IC task ($\text{FC}_i$), yielding a final dimension of $w \times \h_{ij} +  \h({\text{SL}\_{\text{hist}}_i}) +   \h({\text{FC}_i})$. Finally, a Gated Recurrent Unit (GRU) renders the labels auto-regressive followed by \emph{softmax} layer.

\label{sec:experiments_f}

\section{Experiments}

\paragraph{Datasets:} Since there are no existing public datasets for contextual \pretty{IC} and \pretty{SL} task, we use two in-house datasets for evaluation - \textbf{Booking} dataset,\footnote{Dataset will be released to the public} which is a variation of \pretty{DSTC}-2 dataset \cite{williams2014dialog} with intent, slot and dialog act annotations, and \textbf{Cable} dataset, a synthetically created conversational dataset. 
The Booking dataset contains 9,351 training utterances (2,200 conversations) and 6,727 test utterances, with 19 intents and 5 slot types. Cable dataset comprises 1856 training utterances, 1,814 validation utterances and 1,836 test utterances, with 21 intents and 26 slot types.\footnote{Detailed data stats provided in Appendix B}
In addition, we also evaluate the model on non-contextual IC-SL public datasets - \pretty{ATIS} \cite{hemphill90} and \pretty{Snips} \cite{coucke2018snips}.

\paragraph{Experimental setup:} 
To emphasize the importance of contextual signals in modeling, we first devise a non-contextual baseline of our \pretty{Casa-nlu} model, \pretty{Nc-nlu}. It is similar to \pretty{Casa-nlu} in terms of model architecture except no contextual signals are used. 
Since neither datasets nor implementation details were shared in previous works on contextual NLU \cite{Bhargava2013EasyCI, Shi2015ContextualSL}, we also implement another baseline, \pretty{Cgru-nlu} that uses \pretty{GRU} \cite{cho2014learning} instead of self-attention for temporal context fusion. For fair comparison to existing non-contextual baselines, no pre-trained embeddings are used in any of the experiments though the model design can easily benefit from pre-training.

\section{Results and Analysis}
\label{sec:results_f}

Table~\ref{tab:non_context_results} shows the performance of our non-contextual model on two datasets, \pretty{ATIS} and \pretty{Snips}. As shown, we obtain a \textbf{new SOTA} for IC on both the datasets.\footnote{Since we compute token-level F1, SL performance is not compared to results reported in previous work} We hypothesize that the high performance is due to the utterance-level position-aware multi-dimensional self-attention.

As shown in Table \ref{tab:saca_results}, \pretty{Casa-nlu} model outperforms non-contextual NLU on both the Booking and Cable datasets. Further, \pretty{Casa-nlu} model significantly \textbf{outperforms} \pretty{Cgru-nlu} on the Cable dataset by 7.26\% on IC accuracy and 5.31\% on SL F1 absolute, respectively. We believe the reason for strong performance yielded by the \pretty{Casa-nlu} model is due to its multi-dimensional nature, where we learn different weights for different dimensions within the context feature vector ${\mathbf{T}_i}$. This enables the model to learn different attention distributions for different contextual signals leading to more robust modeling compared to \pretty{Cgru-nlu} model. Table \ref{tab:saca_divided_results} gives further breakdown of the results by showing performance on
first
vs. follow-up turns in a dialogue. For the more challenging follow-up turns, \pretty{Casa-nlu} yields significant gains over the baseline IC performance.

\begin{table}
\small
\begin{center}
\setlength\tabcolsep{4.4pt} 
\begin{tabular}{@{}lcccc@{}}
\hline
\bf Model & \multicolumn{1}{@{}c@{}}{\bf \pretty{ATIS}} & \multicolumn{1}{@{}c@{}}{\bf \pretty{Snips}} \\
\hline
  \specialcell[c]{RNN-LSTM*  \cite{hakkani2016multi}}  & 92.6 & 96.9 \\
  \specialcell[c]{Atten.-BiRNN* \cite{joint_ic_sl}} & 91.1 & 96.7\\
  \specialcell[c]{LSTM+attn+gates \cite{Goo2018SlotGatedMF}}  & 94.1 & 97.0 \\
  \specialcell[c]{Capsules Neural Network \cite{zhang2018joint}}  & 95.0 & 97.7 \\
  \hline
  \pretty{Nc-nlu} (ours) & \textbf{95.4} & \textbf{98.4} \\
\hline
\end{tabular}
\end{center}
\caption {Average IC accuracy scores (\%) on non-contextual datasets. *: as reported in \cite{Goo2018SlotGatedMF}}
\label{tab:non_context_results}
\end{table}

\begin{table}
\small
\begin{center}
\setlength\tabcolsep{11.5pt} 
\begin{tabular}{@{}lcccc@{}}
\hline
\bf Model & \multicolumn{2}{@{}c@{}}{\bf Booking} & \multicolumn{2}{@{}c@{}}{\bf Cable} \\
\hline
 & IC & SL & IC & SL \\
\hline
 \pretty{Nc-nlu} & 91.11 & 88.21 & 44.98 & 27.34 \\
  \pretty{Cgru-nlu} & 94.86 & 88.47 & 66.68  & 51.58 \\
  \hline
  \pretty{Casa-nlu} & \textbf{95.16}  & \textbf{88.80}  & \textbf{73.94}  & \textbf{56.97} \\
\hline
\end{tabular}
\end{center}
\caption {IC accuracy and SL F1 scores (\%) for the three models \pretty{Nc-nlu}, \pretty{Cgru-nlu}, \pretty{Casa-nlu} on the 2 contextual datasets - {Booking} and {Cable}.}
\label{tab:saca_results}
\end{table}

\begin{table}[ht]
\small
\begin{center}
\setlength\tabcolsep{11.5pt} 
\begin{tabular}{@{}lcccc@{}}
\hline
\bf Model & \multicolumn{2}{@{}c@{}}{\bf Booking} & \multicolumn{2}{@{}c@{}}{\bf Cable} \\
\hline
& Ft & FU & Ft & FU \\
\hline
  \pretty{Nc-Nlu}& 95.67 & 89.51 & 40.35 & 48.33 \\
  \pretty{Cgru-Nlu} & 98.93 & 94.24 & 46.6  & 72.22 \\
  \hline
  \pretty{Casa-Nlu} & \textbf{99.70}  & \textbf{94.45}  & \textbf{47.44}  & \textbf{81.25} \\
\hline
\end{tabular}
\end{center}
\caption {IC accuracy scores (\%) on first (Ft) and follow-up (FU) turns in contextual datasets - {Booking} and {Cable}.}
\label{tab:saca_divided_results}
\end{table}

\label{sec:qualitative_f}


Table \ref{tab:ablation} shows impact of some of the contextual signals on model performance for the Booking validation dataset.
As expected, contextual signals improve IC and SL performance (Configs - II-V). We observe that adding intent history ($\text{I}_{\text{hist}}$) leads to highest gains in IC accuracy (Config - IV).
At the same time, we see that slot history ($\text{SL}_{\text{hist}}$)  has minimal impact on SL performance for this dataset.  Exhaustive experiments showed that the choice of contextual signals is dependent upon the dataset. Our model facilitates switching these contextual signals on or off easily.


\begin{table}[ht]
\small
\setlength\tabcolsep{4pt} 
\begin{tabular}{lcccccc}
\hline
\textbf{Config} & 
$\text{I}_{\text{hist}}$ & 
$\text{SL}_{\text{hist}}$ &
$\text{Utt}_{\text{hist}}$ &
$\text{DA}_{\text{hist}}$ &
\text{$\text{IC}$} & \text{$\text{SL}$} \\
\hline
I   &x  &x  &x  &x  &89.44  &97.62 \\
II   &x  &x  &\checkmark  &x  &92.52  &98.09 \\
III   &x  &\checkmark  &\checkmark  &\checkmark  &92.35  &98.33 \\
IV   &\checkmark  &\checkmark  &\checkmark  &x  &95.27  &98.34 \\
V   &\checkmark  &\checkmark  &\checkmark  &\checkmark  &96.53  &98.25 \\

\hline
\end{tabular}
\caption {Impact of contextual signals on IC accuracy and SL F1 scores (\%) on Booking validation set for \pretty{Casa-nlu}}
\label{tab:ablation}
\end{table}

\begin{figure}[h]
\centering
\includegraphics[width=0.5\textwidth, height=50pt]{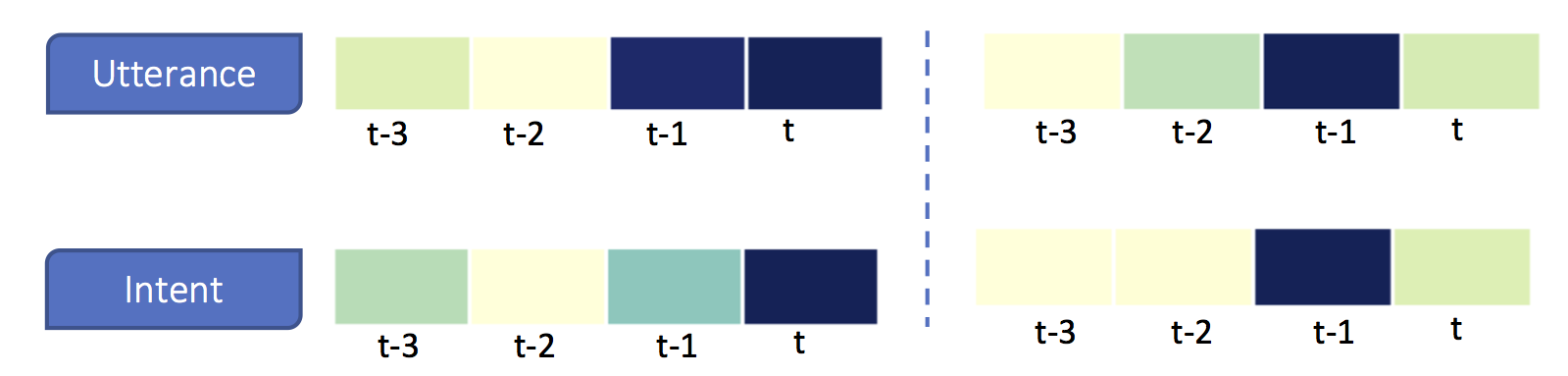}
\caption{\label{attn_1} \small \textbf{Visualization of attention weights} given to two different context signals - previous utterances (top) and previous intents (bottom) for turns t=2 (left) and t=3 (right) from Figure~\ref{tease};
darker colors reflect higher attention weights.}
\end{figure}

\textbf{Qualitative Analysis:} 
Using example conversation in Figure~\ref{tease}, we highlight the relevance of contextual information in making intent predictions
by visualizing attention weights  $\mathbf{P}(\mathbf{T})$ (Equation \ref{eq:attn}) for different contextual signals as shown in Figure \ref{attn_1}.\footnote{For each context signal, attention weights are averaged across all its feature dimensions in $\textbf{P}(\textbf{T})$}
At user turn \#2, the \emph{intent} is switched to \emph{UpgradeService} which the model successfully interprets by paying less attention to previous intents. 
At turn \#3, however, contextual information is critical as user responds to elicitation by agent 
and hence model emphasizes on 
last utterance and intent rather than the current or other previous turns.


\vspace{-10pt}

\section{Conclusion}
\label{sec:conclusion}
We proposed 
\pretty{Casa-nlu} 
model that uses a variable context and various contextual signals in addition to the current utterance to predict the intent and slot labels for the current turn. \pretty{Casa-nlu} achieves gains of over 7\% on IC accuracy for {Cable} dataset over \pretty{Cgru-nlu} baseline, and almost 29\% over non-contextual version. This highlights the importance of using contextual information, meanwhile showing that, learning correct attention is also vital for NLU systems.    


\section{Implementation Details}
\label{sec:implementation_f}

We use hidden layer size of 56 with dropout probability of 0.3. Context history window $K$ was varied from 1 to 5 and the optimal value of 3 was selected.
Word embeddings are trained from scratch using an embedding layer size of 56. Adam \cite{kingma2014adam} algorithm with initial learning rate of 0.01 gave the optimal performance. Concatenation window size $w$ of 3 is used. $\alpha$ and $\beta$ in loss objective are set to 0.9. Early stopping is used with patience of 10 and threshold of 0.5. Each model is trained for 3 seeds and scores averaged across the seeds are reported.
\section{Acknowledgements}
The authors would like to thank the entire AWS Lex Science team for having insightful discussions and providing feedback with experiments. The authors would also like to express their gratitude to Yi Zhang for his generous help and suggestions for this work.




\bibliography{acl2019}
\bibliographystyle{acl_natbib}
\clearpage
\newpage
\begin{appendices}


\section{Utterance Encoding}

We first take the current user utterance for turn $i$, denoted by $\mathbf{U}_i\in \mathbb{R}^{d_e\times k}$ where $k$ represents maximum number of tokens in the utterance.
Then we pass this vector through an embedding layer to generate a hidden state, $\mathbf{H}_i \in\mathbb{R}^{d_h\times k}$ where $d_h$ is the hidden layer dimension. We add learned absolute position embedding, $\mathbf{p}_u \in \mathbb{R}^{d_h\times k}$ \cite{gehring2017convolutional}, resulting in a  new vector, $\mathbf{H}_i = \mathbf{H}_i + \mathbf{p}_u$. The intuition is that this embedding will capture syntactic information.

We pass this input through a \pretty{DiSAN} unit that broadly comprises  token-to-token (t2t) attention, masks, and source-to-token (s2t) attention:
\begin{itemize}
\item On input $\mathbf{H}_i$, we first apply a multi-dimensional \textbf{t2t} attention layer that encodes the dependency between a token and all other tokens in the sentence. In addition, forward and backward masks are added to attention computation to incorporate directional information;
\item For each of these masks, we apply a fusion gate that controls the flow of information from the original hidden representation $\mathbf{H}_i$ and the mask outputs, $\mathbf{Hm}_i^F$ and $\mathbf{Hm}_i^B$, generating contextualized representations $\mathbf{C}_i^F$ and $\mathbf{C}_i^B$, respectively.

$$\mathbf{G} = \sigma(W^{(1)}\mathbf{H}_i + W^{(2)}\mathbf{Hm}_i^F + \mathbf{b})$$
$$\mathbf{C}_i^F = \mathbf{G} \odot \mathbf{H}_i + (1 - \mathbf{G}) \odot \mathbf{Hm}_i^F$$

Similarly, we obtain $\mathbf{C}_i^B$. Both $\mathbf{C}_i^F$ and $\mathbf{C}_i^B$ are concatenated to render $\mathbf{C}_i$;
\item Finally, we have a multi-dimensional \textbf{s2t} attention which learns a gating function that determines, element-wise, how to attend to each individual token of the sentence. It takes $\mathbf{C}_i$ as input and outputs a single vector for the entire sentence.
$$\mathbf{F(C_i)} = W^{(1)T}\sigma(W^{(2)}\mathbf{C}_i + \mathbf{b}) + \mathbf{b}$$
$$\h({\text{Utt}_i}) = \mathbf{F(C_i)} \odot \mathbf{C_i}$$

\end{itemize}


\section{Datasets}

Below is the detailed description of both the conversational datasets:
\begin{enumerate}
    \item \textbf{Booking} provides a shared test framework containing conversations between human and machines for the domain of restaurant. In original \pretty{DSTC-2} data, a user's goal is to find information about restaurants based on certain constraints. The original data contains \textit{states} and \textit{goals} as it is mainly targeted for \pretty{DST} tasks \cite{glad, statenet}. To convert it to \pretty{IC-SL} task, we perform pre-processing on states and goals present in original data to derive intent labels and slots for each user utterance. Some of the sample intents are 'confirm\_pricerange', 'request\_slot\_area', etc. There are only 3 slots - 'food', 'pricerange' and 'area'.
    \item \textbf{Cable} is a synthetic  dataset developed in-house.  It is more diverse compared to {Booking} dataset. It is based on user conversations in the cable service domain. It includes 18 intents like 'ViewDataUsage', 'Help', 'StartService', etc. It has total of 64 slots such as 'UserName', 'CurrentZipCode', etc., yielding a more challenging dataset for modeling.
    \item \textbf{\pretty{Snips}} dataset contains 16K crowdsourced queries. It has total of 7 intents ranging from ‘Play Music’ to ‘Book Restaurant’. Training data has 13,784 utterances and the test data consists of 700 utterances.
    \item \textbf{\pretty{ATIS}} dataset contains 4,978 training utterances and 893 test utterances. There are total of 18 intents and 127 slot labels.
\end{enumerate}

\begin{table}[h]
\centering
\setlength\tabcolsep{1.5pt} 
\begin{tabular}{lcccc}
\textbf{Dataset} & \textbf{Booking} & \textbf{Cable} & \textbf{\pretty{ATIS}} & \textbf{\pretty{Snips}} \\
\hline
Train Size            & 9351                & 1856           & 4978          & 13784          \\
Val Size              & 4691                  & 1814           & --            & --             \\
Test Size             & 6727                & 1836           & 893           & 700            \\
\#Intents      & 19                  & 21             & 18            & 7              \\
\#SL          & 5                   & 26             & 127           & 41             \\
\#DA          & 4                   & 4              & --            & --             \\
\#ConvLen     & 4.25                & 4.68           & --            & --             \\          
\hline
\end{tabular}
\caption{\label{data_table}Data statistics for {Booking}, {Cable}, \pretty{ATIS} and \pretty{Snips}. Legend - \pretty{SL}: Slot Labels; \pretty{DA}: Dialog Acts; ConvLen: Average Conversation length}
\end{table}
\end{appendices}

\end{document}